\documentclass[journal]{elsarticle}
\usepackage[T1]{fontenc}
\usepackage{float}
\usepackage{lineno,hyperref}
\modulolinenumbers[5]

\biboptions{authoryear,sort&compress}

\usepackage{times}  
\usepackage{helvet}  
\usepackage{courier}  
\usepackage{url}  
\usepackage{graphicx}  
\usepackage{subfigure}
\usepackage{colortbl,booktabs}
\usepackage{amsmath}
\usepackage{amssymb}
\usepackage{multirow}
\usepackage{color}
\usepackage{geometry}

\begin{document}

\begin{frontmatter}

\title{Shape and Margin-Aware Lung Nodule Classification in Low-dose CT Images via Soft Activation Mapping}

\author[fudan]{Yiming Lei}

\author[fudan]{Yukun Tian}
\author[rpi]{Hongming Shan}

\author[fudan]{Junping Zhang\corref{correspondingauthor}}
\cortext[correspondingauthor]{Corresponding author}
\ead{jpzhang@fudan.edu.cn}

\author[rpi]{Ge Wang}

\author[harvard]{Mannudeep K. Kalra}

\address[fudan]{Shanghai Key Laboratory of Intelligent Information Processing, School of Computer Science, Fudan University, Shanghai 200433, China}
\address[rpi]{Department of Biomedical Engineering, Rensselaer Polytechnic Institute, Troy, NY 12180 USA}
\address[harvard]{Department of Radiology, Massachusetts General Hospital, Harvard Medical School, Boston, MA 02114 USA}

\begin{abstract}
A number of studies on lung nodule classification lack clinical/biological interpretations of the features extracted by convolutional neural network (CNN). The methods like class activation mapping (CAM) and gradient-based CAM (Grad-CAM) are tailored for interpreting localization and classification tasks while they ignored fine-grained features. Therefore, CAM and Grad-CAM cannot provide optimal interpretation for lung nodule categorization task in low-dose CT images, in that fine-grained pathological clues like discrete and irregular shape and margins of nodules are capable of enhancing sensitivity and specificity of nodule classification with regards to CNN. In this paper, we first develop a soft activation mapping (SAM) to enable fine-grained lung nodule shape \& margin (LNSM) feature analysis with a CNN so that it can access rich discrete features. Secondly, by combining high-level convolutional features with SAM, we further propose a high-level feature enhancement scheme (HESAM) to localize LNSM features. Experiments on the LIDC-IDRI dataset indicate that 1) SAM captures more fine-grained and discrete attention regions than existing methods, 2) HESAM localizes more accurately on LNSM features and obtains the state-of-the-art predictive performance, reducing the false positive rate, and 3) we design and conduct a visually matching experiment which incorporates radiologists study to increase the confidence level of applying our method to clinical diagnosis. 
\end{abstract}


\end{frontmatter}


\section{Introduction}

Lung cancer is one of the most fatal malignant diseases. According to the World Health Organization, there will be ten million deaths from lung cancer by the year 2030 \citep{who_descreption}. To reduce the mortality of lung cancer \citep{in2007, national2011reduced}, several challenges must be addressed including the histologic and genetic heterogeneity of lung cancers and other ailments that often co-exist with lung cancer due to the shared risk factor of smoking such as coronary heart disease and chronic obstructive lung diseases \citep{national2011reduced}. Early detection of lung cancer with low dose CT (LDCT) \citep{CT} has been recently approved in some countries to improve patient survival. However, LDCT has a high false positive rate since many detected pulmonary nodules are benign. Radiologists rely on features such as lesion size, attenuation, margins, growth and stability \citep{AIRadiology2018,2018radiomics} to assess the likelihood of malignancy, and that triggers either follow up LDCT (for less suspicious nodules) or additional diagnostic testing (for more suspicious nodules). In this paper, \textit{we mainly focus on nodule categorization task and LNSM visual analysis rather than nodule detection}. For this purpose, there are two steps in the traditional medical imaging domain using machine learning \citep{2012radiomics}: feature extraction such as using  Local Binary Pattern (LBP) \citep{LBP}, and then nodule classification such as via support vector machine (SVM) \citep{LEI2017281}. Unlike deep learning, a major disadvantage of this scheme is its limited ability in representing various lung nodule shapes and margins.

\begin{figure}
 \centering
   \setlength{\abovecaptionskip}{0pt}
    \setlength{\belowcaptionskip}{5pt}
     \subfigure{
      \includegraphics[width = 5in]{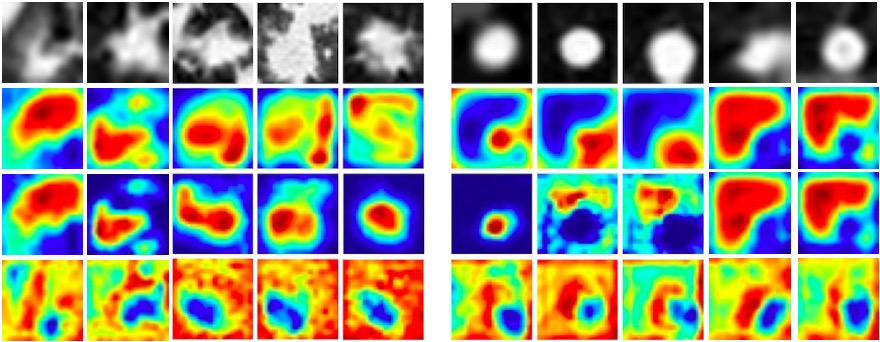}
      }
 \caption{Visualization of LNSM features. The first row illustrates different types of nodule ROIs, the second row shows the CAM \citep{Zhou_2016_CVPR}, the third row shows Grad-CAM \citep{gradcam} and the fourth row is our HESAM maps. The three left columns represent malignant nodules, and the three right columns display benign nodules.}
 \label{cam_grad_hesam}
\end{figure}

Over the past decade, deep learning has made tremendous progress in image classification. And it is well-known that object shape is the single most important cue for human object recognition \citep{1988shape} as well as for deep neural networks \citep{humanshape}. During the learning and decision processes, CNNs often exhibit shape bias, which is related to human cognitive psychology \citep{2017cognitive}, i.e., shape plays a dominant role in recognition behavior over color and texture. Based on this observation, the goal of proposed methods can be expressed as improving lung nodule classification performance and simultaneously providing the attention regions (LNSM) as the decision clues. Moreover, we expect these clues to coincide well with ``where the radiologists care about''.

While acquiring attention regions in lung nodule ROIs (region of interests) is different from localizing objects of interest in natural images. In the tasks of classification and detection on ImageNet or COCO \citep{ILSVRC, coco}, the deep models learn to recognize one or more salient objects in one image. CAM \citep{Zhou_2016_CVPR} and Grad-CAM \citep{gradcam} are two popular approaches for interpreting classification of natural images based on global average pooling (GAP) and gradients of the class-specific score with respect to all pixels in final feature maps, respectively. However, both of them ignore fine-grained local features, and intend to attach attention to salient nodule bodies especially for benign nodules (Fig. \ref{cam_grad_hesam}).

In contrast to natural images, lung nodule ROIs contain only one nodule and a totally different background. Furthermore, according to \citep{hoo2016deep, maligVSben}, 1) if only performing a coarse feature analysis in malignant and benign nodule regions, the performance of the predictive system will degenerate; 2) to improve the lung nodule classification, it is necessary to perform a fine-grained feature analysis so that discriminative lesion descriptors such as ground glass (GG), cavity (CA), micronodules (MN) and consolidation (CD) can be perceived by CNNs. To achieve this goal, we assume that if a classifier focuses more on LNSM features rather than salient nodule bodies, it may perform better in classification since it is in correspondence with human cognitive psychology \citep{2017cognitive}. This motivates us to endow the CNNs with the capability of being sensitive to the regions with fine-grained LNSM features, so we propose soft activation mapping (SAM) to discretize the final Conv feature maps (Fig. \ref{sam}). Then the importance of each final feature map is obtained via average pooling followed by a fully connected layer. Thence, the importance of one final feature map is learned by taking every local region of this feature map into consideration. This contribution enables the model to refine the fine-grained local regions (LNSM). On this basis, a high-level enhanced SAM (HESAM) is proposed to re-utilize high-level features with the lowest resolution in an encode-decode structure to improve classification accuracy and rectify LNSM localization \citep{shan2019nature}. The main difference between proposed SAM and CAM is that the final feature maps are split into local patches which contribute independently to final importance scores. Our experimental results show that SAM indeed focuses on relatively discrete and irregular features around the nodule bodies and HESAM further improves classification and localization performance.

Another issue in machine learning-based lung LDCT image analysis is that insufficient labeled data is available for training because of the involved high annotation cost. A number of studies \citep{fully3d2017,multiscale_cnn,2016multiview,2017tumornet} were conducted to automate this task to various degrees. However, none of the methods is totally satisfactory in extracting features from a whole CT volume, which contains diverse features specific to the categorization of nodules and individual characteristics. In this paper, we propose to take volume images of lung nodules into a 2D network for streamlined processing \citep{havaei2017brain}. Experiments show that compared with other recently published methods, our method achieves a state-of-the-art classification accuracy.

Main contributions are summarized as follows:
\begin{itemize}
  \item SAM is introduced to localize LNSM in fine-grained regions, facilitating lung nodule classification;
  \item HESAM is proposed to combine SAMs and high-level features to achieve the best performance of classification and LNSM localization;
  \item With the input of multiple CT slices, our model alleviates overfitting without conventional data augmentation, thereby improving the predictive performance.
\end{itemize}

\section{Related Work}

In this section, we will give a brief survey on following three aspects: 1) CNNs applied to medical image classification \citep{zhang2019medical,LDCTcnn,litjens2017survey,schlemper2019attention}, 2) UNet and its variants, 3) two activation mapping methods for attention regions localization.

\textbf{CNNs-based medical image classification.} One difficulty in lung nodule classification is the limited number of annotated CT slices. As in most deep learning studies, Shan et al. augmented the set of lung nodule slices through rotation, flip and shift, etc \citep{fully3d2017}. However, these simple operations are not effective enough for alleviating overfitting. Setio et al. extracted 2D patches from nine symmetrical planes of a cube where a potential candidate locates at the center of the corresponding patches \citep{2016multiview}. TumorNet \citep{2017tumornet} obtained three candidates of one patch by projecting median intensity along three coordinate axes. Nevertheless, the features learned by these methods cannot cover the lesions over all slices. Naturally, researchers intended to use the volume data and 3D networks for achieving better performance. Shen et al. proposed multi-scale CNNs to identify the type of lung nodules \citep{multiscale_cnn}. They simply utilized a model with two convolutional layers and each of them followed by a pooling layer. Wu et al. fed the volume data into an encode-decode structure and accomplished the tasks including reconstruction and classification \citep{pn-samp}. DeepLung \citep{deeplung} firstly applied 3D dual path network (DPN) \citep{dpn} to detect and classify lung nodule simultaneously. Although some 3D models achieve considerable results in biomedical engineering, they are still difficult to perform better due to large-scale parameters that are to be learned \citep{deeplung,winkels2019pulmonary}.

\textbf{Variants of UNet.} UNet was proposed to conduct the medical image segmentation with the skip connection \citep{unet}, which can convey the early features and combine them with features of later layers \citep{shan20183d, densenet} so that high-resolution features can be preserved. Dou et al. extracted volume candidates with various nodule diameters \citep{2017multilevel3d}. These 3D CNNs learn more spatial information from different slices. An advantage is that the 3D network can retrace the success of the 2D network if a large amount of labeled data are available \citep{3d_retrace_2d}. Furthermore, 3D-UNet \citep{3d-unet,pn-samp} was proposed to segment volumetric medical data and conduct multi-task learning, by performing classification after the output layer of 3D-UNet.

\textbf{CAM: Class activation mapping via global average pooling.}
Instead of adding FC layers on top of the feature maps, GAP was proposed as a network in network and fed the resultant vectors directly into the softmax layer \citep{NIN}. Furthermore, GAP is the key component for attention maps generation through conducting GAP on the final feature maps, followed by one FC layer \citep{Zhou_2016_CVPR}. The computation of the class activation map can be written as follows:
\begin{align}
  &S_{c} = \sum_{k} \omega_{k}^{c} \sum_{x,y} f_{k}(x,y) = \sum_{x,y}\sum_{k} \omega_{k}^{c} f_{k}(x,y),\\
  &M_{c}(x,y) = \sum_{k} \omega_{k}^{c} f_{k}(x,y),
\setlength\belowdisplayskip{0.5pt}
\end{align}
where $F_{k} = \sum_{x,y} f_{k}(x,y)$ and $f_{k}(x,y)$ are results of performing GAP on unit $k$ and the activation of the last convolutional layer at unit $k$ respectively. $S_{c}$ is used as the input to the softmax. $\omega_{k}^{c}$ represents the importance of $F_{k}$ for class $c$ \citep{Zhou_2016_CVPR}. This method can identify the extent of an object, and make sense for natural object localization. In order to obtain more fine-grained features or more precise object localization, Grad-CAM is proposed to shrink the localization to a smaller range.

\textbf{Grad-CAM: visual explanations via Gradient-based localization.} In \citep{gradcam}, the authors pointed out that a good visual explanation from the model justifying a predicted class should be class discriminative (i.e. localize the target category in the image) and high-resolution (i.e. capture fine-grained details). To avoid the drawback of CAM that is unable to capture fine-grained details, Grad-CAM uses the gradient information flowing into the last convolutional layer of CNN to obtain the importance of \emph{each neuron} for a decision of interest. The weight $\alpha_{k}^{c}$ is defined as follows:
\begin{align}
  \alpha_{k}^{c} = \underbrace{ \frac{1}{Z} \sum_{i} \sum_{j} }_{GAP} \frac{\partial{y^{c}}}{\partial{A_{i,j}^{k}}},
\end{align}
where $A^{k}$ represents the $k$th feature map of a convolutional layer, $Z$ is the number of pixels in $A^{k}$, and $y^{c}$ denotes the score for class $c$. An advantage of Grad-CAM is that it requires no changing of model structures and no re-training. In addition, Grad-CAM applied a ReLU to the linear combination of maps, as described in Eq. \eqref{relu_grad_cam}:
\begin{align}
  L_{Grad-CAM}^{c} = ReLU(\sum_{k} \alpha_{k}^{c} A^{k}).
  \label{relu_grad_cam}
\end{align}
ReLU makes the model be interested in those features that have a positive influence on the class of interest. Negative pixels that are likely to belong to other categories in the image can be filtered out by ReLU.

In short, CAM and Grad-CAM focus more on the class-specific objects (e.g., nodule bodies which can be seen in Fig. \ref{cam_grad_sam}, Fig. \ref{hesam} and \textit{Fig. A.2 in Appendix A}). However, they are inferior in recognizing local LNSM features around the nodule bodies. To address this issue, we propose a new component consisting of average pooling and the FC layer following each final feature map, which enables the model to attach attention to margins of nodules rather than nodule bodies.

\section{Methods}
\begin{figure}[H]
  \centering
  \setlength{\abovecaptionskip}{2pt}
  \subfigure{
    \includegraphics[width = 6in]{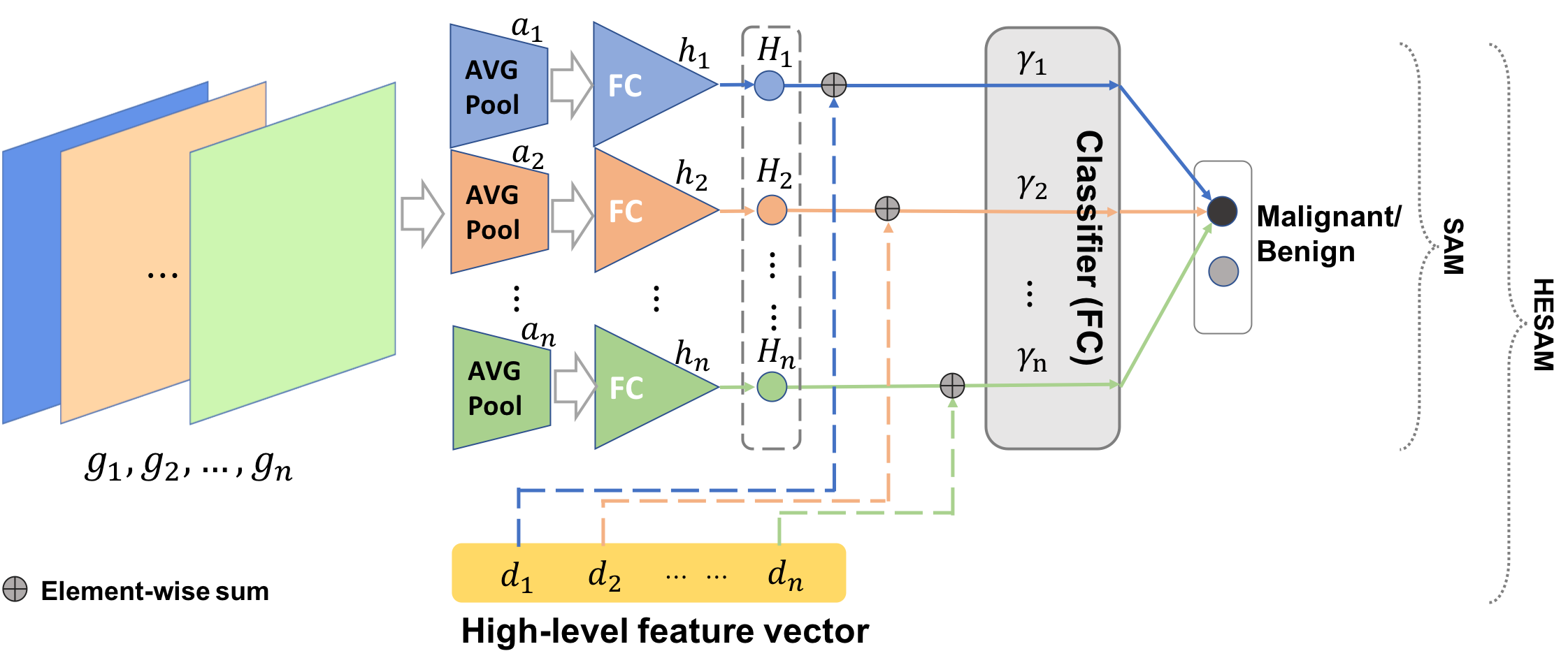}
  }
  \caption{Soft activation mapping (SAM) and its high-level feature enchancement scheme (HESAM). $g_{1}, g_{2}, ..., g_{n}$ represent final Conv feature maps, $\gamma_{1}, \gamma_{2}, ..., \gamma_{n}$ are the weights of the classifier. Each FC layer $h_{k}$ (colored triangle) maps the AVG pooling result of corresponding final Conv feature into one single value $H_{k}$. Note that $k\in{1, ..., n}$. $d_{k}$ represents the high-level feature which is obtained by UNet in this paper (Implementation Details, {\it SubSection 3.2}).}
  \label{sam}
\end{figure}

In this section, we describe our soft activation mapping (SAM) part in Fig. \ref{sam} and implementation details in Fig. \ref{net}. 

\subsection{Soft Activation Mapping (SAM)}
\begin{figure}
  \centering
  \subfigure{
    \includegraphics[width = 6in, height = 2.6in]{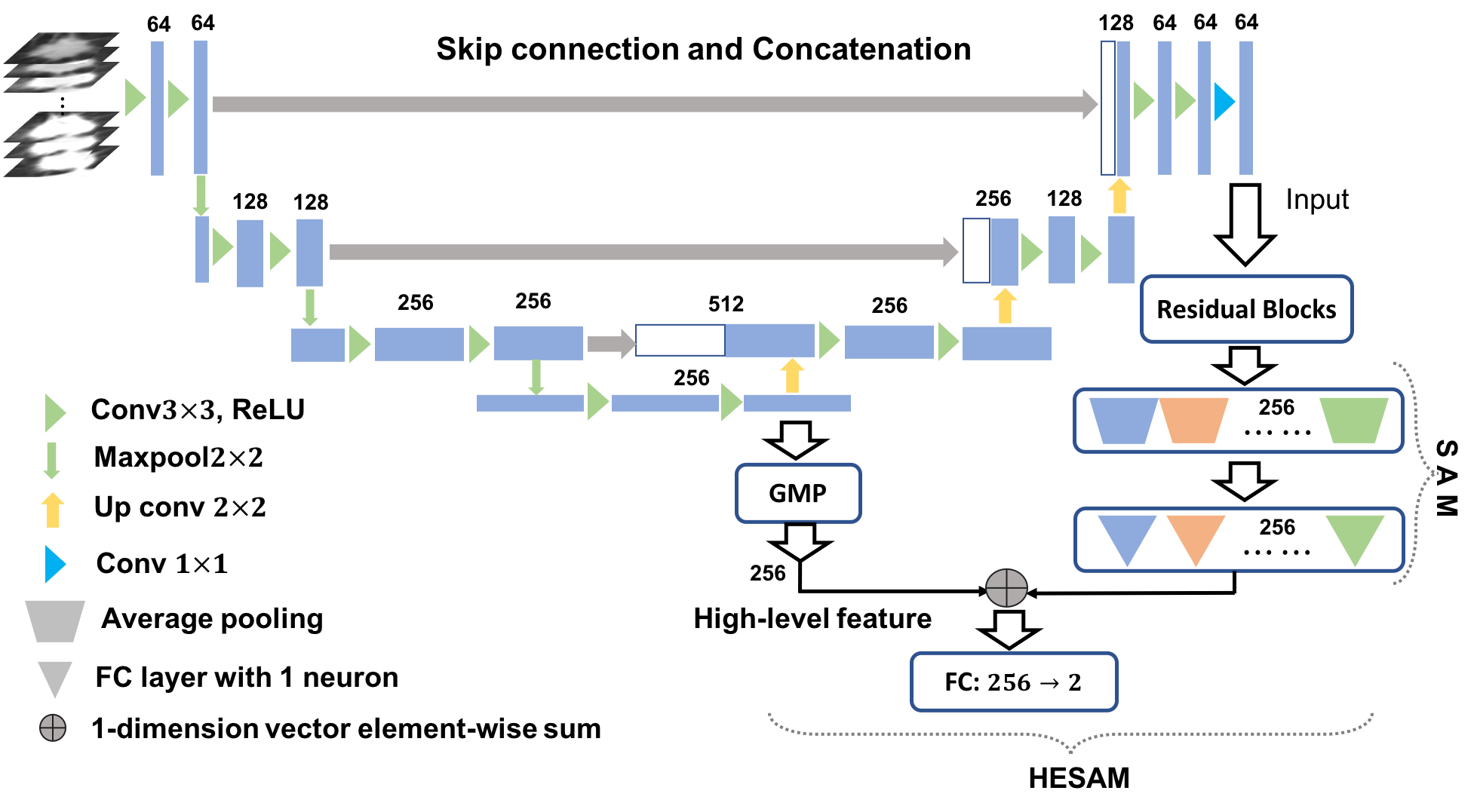}
  }
  \setlength{\abovecaptionskip}{0pt}
  \caption{Implementation of proposed SAM and HESAM. All the Conv layers in the residual blocks followed by BN \citep{BN} and ReLU. GMP and FC represents global maxpooling and fully-connected layers respectively. Average pooling setting: kernel\_size = 5, stride = 2, padding = 0.}
  \vspace{0.1cm}
  \label{net}
 \end{figure}

CAM \citep{Zhou_2016_CVPR} has verified the power of GAP in object localization. For some nodule features at local or gracile regions, however, GAP will lose its merit. Therefore, we proposed a soft activation mapping, as illustrated in Fig. \ref{sam}, to address this issue. It is defined as follows:
\begin{align}
  A^{k}(x', y') &= a(g_{k}(x,y)), k \in \{1,2,...,n\}, \\
  H_{k} &= h_{k}(A^{k}),
\end{align}

where $g_{k}(x,y)$ represents the activation function of unit $k$ in the last Conv layer (i.e., the last Conv layer in the residual blocks (Fig. \ref{net}) in our implementation) at a spatial location $(x,y)$, $g_{k}$ is the $k$th feature map, $a$ is the average (AVG) pooling layer, and $A^{k}(x', y')$ represents the result generated by AVG pooling at the spatial location $(x,y)$ of $g_{k}$. In this study, we call $A^{k}$ (features after the final AVG pooling layer) the \emph{minor feature}, where the value at $(x',y')$ is mapped from the region centered by $(x,y)$ in final Conv features. $h_{k}$ denotes the FC layer which follows $A^{k}$, and each $h_{k}$ has only one neuron, i.e., every $H_{k}$ is a single value. Therefore, the final $n$ feature maps are projected into $n$ values.
The input to the softmax, for a given class $c$, is $S'_{c}$. The soft activation map $M'_{c}$ of class relevance is defined as follows:
\begin{align}
  M'_{c} &=\sum_{k} \gamma_{k}^{c} g_{k}(x,y), \\
  S'_{c} &= \sum_{k} \gamma_{k}^{c} H_{k} \\
         &= \sum_{k} \gamma_{k}^{c} \underbrace{ \sum_{\ell} \beta_{\ell} A_{\ell}^{k}(x', y') }_{Fully Connected} \\
         &= \sum_{k} \gamma_{k}^{c} \sum_{\ell} \beta_{\ell} a_{\ell}(g_{k}(x, y)).
\end{align}
Finally, the output of the softmax for class $c$ is given by $\frac{exp(S'_{c})}{\sum_{c} exp(S'_{c})}$. $M'_{c}$ indicates the importance of activation at a spatial location $(x,y)$ for classifying an image of class $c$. $\gamma_{k}^{c}$ represents the importance of $g_{k}$ for class $c$. $\beta_{\ell}$ denotes the parameters of FC layer which is following the $k$th minor feature, and $A_{\ell}^{k}(x', y')$ represents the $\ell$th value in minor feature $A^{k}$.

Note that CAMs \citep{Zhou_2016_CVPR}, Grad-CAMs \citep{gradcam} and SAMs all assume that the final Conv feature maps contain more valuable and more semantic information about the corresponding input image. Each feature map matches a certain interested part of that class. Then, the linear combination of these features is performed with respect to a group of weights ($\gamma_{k},k\in\{1,2,...,n\}$ in Eq. (7)), i.e., each final Conv feature contributes differently to the final localization. Nevertheless, the main distinction among these three methods lies in \emph{how to compute the weights related to the corresponding features}. CAM transforms one feature map into one single averaged value over all pixels. This approach avoids low activations that reduce the output of the particular maps and high activations that augment the output of the particular maps, which may happen with global max-pooling (GMP). Grad-CAM obtains the more accurate localization of interests than CAM. However, for the specificity of lung nodules, more variations of margin and shape morphology around the nodule bodies are closely related to categorization of a nodule \citep{maligVSben}. That is to say, these variations belong to one class (e.g., malignant) since the model makes predictions based on objects' shape \citep{2017cognitive}. So, localizing attention regions of discrete and various changing patterns in lung nodule ROIs cannot be easily resolved by CAM or Grad-CAM that capture the extent of regions. 

In our SAM, we do not directly project feature maps into their respective single values. By average pooling, some high activations that reflect discriminative features can be averaged just in small receptive fields. Minor features $A^{k}$s maintain in regions of high/low activations. The subsequent FC layers project the local features into single values that will be fed into the final classifier. That is to say, $\gamma_{k}^{c}$ and $\beta_{\ell}$ are optimized simultaneously. In a micro view, $\beta$ acts as the importance of each element in minor features $A^{k}$s and each element represents a local part of that final feature map. Hence, each local part of a final feature map impacts the final weight differently, and high/low activations can make contributions through learning process.  Moreover, the magnitude of parameters of SAM is reduced by adding a number of FC layers to each final feature map respectively (\textit{Fig. C.4 in Appendix C}). Besides the parameter reduction, adding one standard FC layer with the same number of neurons as the final feature maps will mix all the information from these maps. For the binary lung nodule classification task, the discriminative and fine-grained information is limited compared with natural images. So, if the number of training samples ($N$) is fixed and small, the standard FC counterpart with a huge amount of parameters is prone to overfitting the training data because of the high model complexity ($P/N$, $P$ denotes the number of model parameters) \citep{relation,understanding}. Further, the effectiveness of adding one standard FC layer is far away from our expectation that local parts of final features can contribute to the final decisions independently (\textit{Appendix C}).

\textbf{High-level enhanced SAMs (HESAMs)}. 
Feature maps from the final Conv layers of the residual blocks cover most of the attention regions (LNSM) since they are in larger size. SAM maintains semantic features according to high/low activations. However, some regions corresponding to middle activations of two categories will mislead the classifier. In order to avoid this problem and improve the performance of our model in SAM version, we add high-level representations to enhance the feature vector generated by SAM (HESAM). The soft activation maps and the input to the softmax are defined as follows:
\begin{align}
   M_{c}^{'h} &= \sum_{k} \gamma_{k}^{ch} g_{k}(x,y), \\
   S_{c}^{'h} &= \sum_{k} (H_{k} + d_{k}) \gamma_{k}^{ch},
\end{align}
where $d_{k}, k\in\{1,2,...,n\}$ are representations of high-level features generated by GMP (Fig. \ref{net}). $d_{k}$ has less direct relationship with class information but only relates to the specific input. That is to say, we will combine the information of salient nodule structures with the output of SAM. Here, $\gamma_{k}^{ch}$s denote the parameters of classifier which is fed with ($H_{k} + d_{k}$). This is different from $\gamma_{k}^{c}$s (Eq. 7) that represent the parameters of classifier taking $H_{k}$ as input. Therefore, $d_{k}$s do not impact fully on the weights $\gamma$ in that they only strengthen the input of classifier at certain dimensions. In other words, some structural-related information, such as salient bodies for benign nodules and irregular shape/margins for malignant nodules, act as an auxiliary enhancement for class specific representations.

\subsection{Implementation Details} 
In order to testify the effectiveness of SAM, we used UNet-based encode-decode structure. It begins with the input of multiple slices, and details of data preprocessing will be described in the Data Preparation section. In the {\bf UNet structure}, we apply three down-sampling blocks and three up-sampling blocks to acquire a representative code of the input. This code is used to enhance the localization ability and classification performance. Similarly, the related settings (kernel sizes, number of neurons, strides, and paddings) of the corresponding layers remain the same as those in \citep{unet}. At the output layer of UNet, there is a convolutional (Conv) layer with kernel size 1. Then, the final output is the segmented image of 1 channel. We specify the number of final $1 \times 1$ kernels as $64$ and feed the resulted 64 feature maps into another residual learning network \citep{ResNet}. {\bf Residual blocks} in Fig. \ref{net} contain two basic blocks with a kernel size of $3$ for all the Conv layers \citep{ResNet}, and these blocks are used to extract more shape and margin features from 64-channel output of UNet structure.

Among most of classification the tasks accomplished by deep convolutional networks, high-level features fed into classifiers will have a great impact on the final performance since they reflect semantic and discriminative information from CT slices. To utilize these features, we add a {\bf Global MaxPooling (GMP)} layer after the end of the third down-sampling block which contains $256$ feature maps of size $4\times4$. Hence, the output of the GMP layer is a $1$ dimensional vector consisting of $256$ elements. Then $4\times4$ high-level feature vector can be decoded to approximate the original image by combining early features through the skip connection \citep{unet}. Therefore, the $256$ dimensional vector can be denoted as an effective representation of the input image in a low dimensional space. Normally, this representation is obtained in the reconstruction task. Then, the parameters are updated based on the reconstruction loss, e.g., Mean Square Error and Kullback-Leibler divergence, and the loss value is computed pixel-wisely. Hence, an input image can be projected as a code into a lower-dimensional subspace, and the code is of high representative ability for the input. In contrast, our implementation uses a classification loss, e.g., cross-entropy loss. This differs from the aforementioned reconstruction task, in which our parameters are updated based on class information. In other words, the code learned by the reconstruction loss tends to represent the input itself, and meanwhile, the classification loss lets the code to be mostly related to features of the input. Therefore, the final down-sampling code is actually the effective representation of features of the input from a certain class. This feature related code is used to locate the LNSM regions precisely, which will be further discussed in experiments. The proposed {\bf SAM} method is attached to the end of the residual blocks, which aims to split patches of final Conv feature maps of residual network. {\bf HESAM} in Fig. \ref{net} is composed of high-level feature enhancement branch and SAM. Note that the output of GMP has the same dimensions with the output of SAM, then the sum of these two outputs is regarded as the input of the final FC layer.

If we want SAM to provide forceful clues of categorization, it must obey the following rule. As stated in \citep{Zhou_2016_CVPR}, the final Conv features for computing CAMs should be in larger size such as $13\times13$ or $14\times14$. Thus, the output feature maps of the residual blocks in our model are of size $16\times16$. Through the AvgPooling layer, the size of each feature map was converted to $6\times6$. After the FC layers with respect to each $6\times6$ feature vector, we obtained a $256$ dimensional vector. The sum of this vector and output of GMP in high-level enhancement phase provides a satisfactory representation of an input in terms of semantics and spatial structure.

\section{Experiments}
In this section, we evaluate the performance of our model, SAM and HESAM on LIDC-IDRI \citep{LIDC} dataset. Experiments show that our proposed model achieves higher classification accuracy and more accurate LNSM region localization compared with several state-of-the-art models. 

\subsection{Data Preparation and Experimental Setting}
LIDC-IDRI is a publicly available dataset of pulmonary nodules from 1010 patients who underwent chest CT LIDC. Each nodule was rated from 1 to 5 by four experienced thoracic radiologists, indicating an increasing probability of malignancy. In this study, the ROI of each nodule was obtained along with its annotated center in accordance with the nodule report, with a square shape of a doubled equivalent diameter. An average score of a nodule was used for assigning probability of malignant etiology. Nodules with an average score higher than 3 were labeled as malignant and lower than 3 are labeled as benign. Some nodules were removed from the experiments in the case of the averaged malignancy score 3, ambiguous IDs, and being rated by only one or two radiologists. All volumes were resampled to have $1 mm$ spacing (original spacing ranged from $0.6mm$ to $3.0mm$) in the z-dimension with $512\times512$ matrices. Finally, the dataset consists of 635 benign and 510 malignant nodules. All experiments are conducted through 5-fold cross-validation. 

In addition, our model requires the input with $Channel \in \{1, 3, 11, 21\}$, which is symmetrically selected on both sides of annotated center, e.g., $Channel = 11$ represents that there are 5 more channels on both sides of the center. Note that $Channel = 1$ denotes the ROI in center slice only. Therefore, we construct 4 datasets (Tab. \ref{dataset}) to evaluate how the performance of deep networks can be impacted by different input channels of nodule volume. As illustrated in Tab. \ref{dataset}, image resolution of all patches is $32\times32$. The $D_{3C}$ dataset comprises patches with the square shape of $p\times diameter, p\in\{2,3,4\}$ around the annotated center, and each input sample consists of 3 patches corresponding to 3 channels. 

The hyper-parameters for all experiments are set as follows: learning rate is $0.0005$ and is adjusted by $1/t, t=10$ every $30$ epochs; mini-batch size is $32$, and weight decay is $0.0001$. The loss function $L$ is cross-entropy loss \citep{crossentropy,zhang2019medical,pesce2019learning}:
\begin{align}
	L(\theta) = y\log\hat{y} + (1-y)\log(1-\hat{y})
\end{align}
where $\hat{y}$ is the label of an input, $y$ represents the prediction obtained by forward computation and $\theta$ is the set of parameters that are to be learned. And optimizer is stochastic gradient decent (SGD) \citep{sgd}. All of our experiments are implemented in the PyTorch \citep{pytorch} and trained with 2 NVIDIA GTX 1080 Ti GPUs.
\begin{table*}
  \centering
  \setlength{\belowcaptionskip}{5pt}
  \caption{Summary of constructed datasets with different channels: $NumOfImages \times Channel \times Height \times Width$}

    \begin{tabular}{p{4cm}p{2.5cm}<{\centering}p{2.5cm}<{\centering}}
      \hline
       Dataset & Train & Test \\
      \hline
       $D_{1C}$ & $916 \times 1 \times 32 \times 32$ & $229 \times 1 \times 32 \times 32$ \\
       $D_{3C}$\citep{multiscale_cnn} & $916 \times 3 \times 32 \times 32$ & $229 \times 3 \times 32 \times 32$ \\
       $D_{11C}$ & $916 \times 11 \times 32 \times 32$ & $229 \times 11 \times 32 \times 32$ \\
       $D_{21C}$ & $916 \times 21 \times 32 \times 32$ & $229 \times 21 \times 32 \times 32$ \\
      \hline
    \end{tabular}
  \label{dataset}
\end{table*}

\subsection{Nodule Classification Using Volume Data}
In this section, we firstly conduct classification experiments using these four mentioned datasets, and the input channel for all models varies from the related dataset. For example, the input channel is set to be 11 for all models when we use $D_{11C}$ dataset. Obviously, this will increase the number of parameters at $1\emph{st}$ Conv layer compared with using $D_{1C}$ and $D_{3C}$ datasets. 

\begin{table*}
  \centering
  \caption{Classification accuracies (\%) with standard deviations on four datasets. Values of sensitivity and specificity are obtained based on $D_{11C}$ dataset. $\centerdot-$CAM and $\centerdot-$SAM denote that corresponding networks performed the following modifications to generate larger feature maps ($16\times16$) at the last Conv layer. ResNet-CAM/SAM: reduce the last pooling layer and strides in each residual block are set as 1; VGG16-CAM/SAM: delete the last $4$ pooling layers and the number of FC layers is reduced to $1$; DenseNet-CAM/SAM: delete the final pooling layer and strides of pooling in transition block are reduced to $1$. SAG-C is the simplified attention gate classification scheme (we only preserved the first Conv layer at each scale in order to alleviate performance collapse in our task) \citep{schlemper2019attention}.}
    \begin{tabular}{p{3cm}p{1.7cm}<{\centering}p{1.7cm}<{\centering}p{1.7cm}<{\centering}p{1.7cm}<{\centering}p{1.3cm}<{\centering}p{1.3cm}<{\centering}}
      \hline
      Method & $D_{1C}$ & $D_{3C}$ & $D_{11C}$ & $D_{21C}$ & Sensitivity & Specificity \\
      \hline
      PN-SAMP  & 84.28$\pm0.006$ & 91.70$\pm0.005$ & 97.82$\pm0.006$ & 97.38$\pm0.006$ & 0.9509 & 0.9763 \\
      SAG-C  & 83.41$\pm0.005$ & 91.70$\pm0.005$ & 96.94$\pm0.004$ & 96.07$\pm0.005$ & 0.9705 & 0.9685 \\
      ResNet18  & 81.22$\pm0.005$ & 91.27$\pm0.005$ & 95.63$\pm0.004$ & 95.79$\pm0.006$ & 0.9411 & 0.9685 \\
      ResNet34  & 80.79$\pm0.005$ & 90.83$\pm0.006$ & 96.07$\pm0.005$ & 95.20$\pm0.006$ & 0.9705 & 0.9763 \\
      VGG16  & \textcolor[rgb]{1,0,0}{85.59$\pm0.006$} & 91.70$\pm0.005$ & 96.69$\pm0.005$ & 96.07$\pm0.007$ & 0.9411 & 0.9763 \\
      DenseNet121  & 79.91$\pm0.007$ & 87.77$\pm0.006$ & 93.45$\pm0.004$ & 90.83$\pm0.006$ & 0.9215 & 0.9527 \\
      \hline
      ResNet18-CAM & 81.66$\pm0.005$ & 88.21$\pm0.005$ & 95.36$\pm0.003$ & 95.20$\pm0.006$ & 0.9216 & 0.9764  \\
      ResNet34-CAM & 81.66$\pm0.006$ & 86.90$\pm0.005$ & 97.82$\pm0.005$ & 96.94$\pm0.006$ & 0.9509 & 0.9781 \\
      VGG16-CAM & 83.41$\pm0.007$ & 89.96$\pm0.005$ & 92.14$\pm0.005$ & 91.70$\pm0.006$ & 0.9117 & 0.9448 \\
      DenseNet121-CAM & 80.35$\pm0.007$ & 86.46$\pm0.006$ & 95.63$\pm0.004$ & 89.08$\pm0.005$ & 0.9313 & 0.9527 \\
      \hline
      ResNet18-SAM & 84.28$\pm0.005$ & 89.96$\pm0.005$ & 97.38$\pm0.004$ & 96.94$\pm0.005$ & 0.9607 & 0.9763 \\
      ResNet34-SAM & 83.41$\pm0.005$ & 88.69$\pm0.006$ & 98.25$\pm0.005$ & 96.51$\pm0.005$ & 0.9509 & 0.9843 \\
      VGG16-SAM & 83.41$\pm0.007$ & 90.83$\pm0.005$ & 94.32$\pm0.006$ & 93.89$\pm0.006$ & 0.9411 & 0.9448 \\
      DenseNet121-SAM & 81.22$\pm0.007$ & 88.65$\pm0.007$ & 97.28$\pm0.005$ & 94.76$\pm0.005$ & 0.9411 & 0.9685 \\
      \hline
      \textcolor[rgb]{0,0,1}{Baseline} & \textcolor[rgb]{0,0,1}{81.66$\pm0.006$} & \textcolor[rgb]{0,0,1}{85.15$\pm0.005$} & \textcolor[rgb]{0,0,1}{94.76$\pm0.004$} & \textcolor[rgb]{0,0,1}{93.01$\pm0.006$} & \textcolor[rgb]{0,0,1}{0.9215} & \textcolor[rgb]{0,0,1}{0.9685} \\
      Ours-CAM & 80.35$\pm0.006$ & 86.46$\pm0.005$ & 96.51$\pm0.004$ & 95.20$\pm0.006$ & 0.9509 & 0.9737 \\
      Ours-SAM & 81.66$\pm0.005$ & 89.08$\pm0.004$ & 98.25$\pm0.004$ & 97.38$\pm0.007$ & 0.9509 & 0.9843 \\
      Ours-HESAM & 83.41$\pm0.005$ & \textcolor[rgb]{1,0,0}{92.58$\pm0.005$} & \textcolor[rgb]{1,0,0}{99.13$\pm0.003$} & \textcolor[rgb]{1,0,0}{98.69$\pm0.006$} & \textcolor[rgb]{1,0,0}{0.9705} & \textcolor[rgb]{1,0,0}{0.9921} \\ 
      \hline
    \end{tabular}
    \label{results_1}
\end{table*}

\textbf{Classification performance.} All the models in our experiments are trained using $4$ datasets. As illustrated in Tab. \ref{results_1}, for all methods, accuracies of using $D_{11C}$ and $D_{21C}$ are higher than those of using $D_{1C}$ and $D_{3C}$, and our method achieves higher specificity (from 0.5\% to 5\%) than others. This demonstrated that various features in multiple slices enhanced the representation of a nodule volume. Actually, the increase of channels reflects that abundant information of lesion features are learned at the same time, i.e., at first convolution layer, each filter convolves an input at transverse plane across all channels. Then, the output of the $1 st$ Conv layer is the aggregation of results with respect to all channels. In the case of $D_{1C}$, features extracted by the first Conv layer have less structural diversity than those of other 3 datasets. $D_{3C}$ fuses the multi-scale information into feature learning. Although maps of $3$ channels of input are based on one slice, the edge variations of a nodule in other slices are ignored. $D_{11C}$ and $D_{21C}$ integrated both shape variations and multi-scale information, thereby the classification results outperform the other two datasets. We can observe that accuracies on $D_{21C}$ are slightly lower than those of $D_{11C}$ for all methods, and this is probably because more channels of slices which are far from center slice contain more similar background information. The performance degradation appears when channel is equal to 3. One possible reason is that the representability of features obtained by both CAM and SAM is weaker than those extracted by corresponding normal models because of the detachment of some additional layers (e.g., pooling or FC layer) \citep{Zhou_2016_CVPR}. Besides this, each channel contains the nodule ROI at different scales while all of them hold the same nodule structure. It is undoubted that there will be more redundant information in the feature maps with larger size. Compared with one channel, this phenomenon merely happens sporadically. In addition to performance degradation, an interesting phenomenon is that models of SAM version perform better than their CAM counterparts.

\begin{figure*}[ht]
  \centering
  \subfigure{
  	\includegraphics[width = 6in]{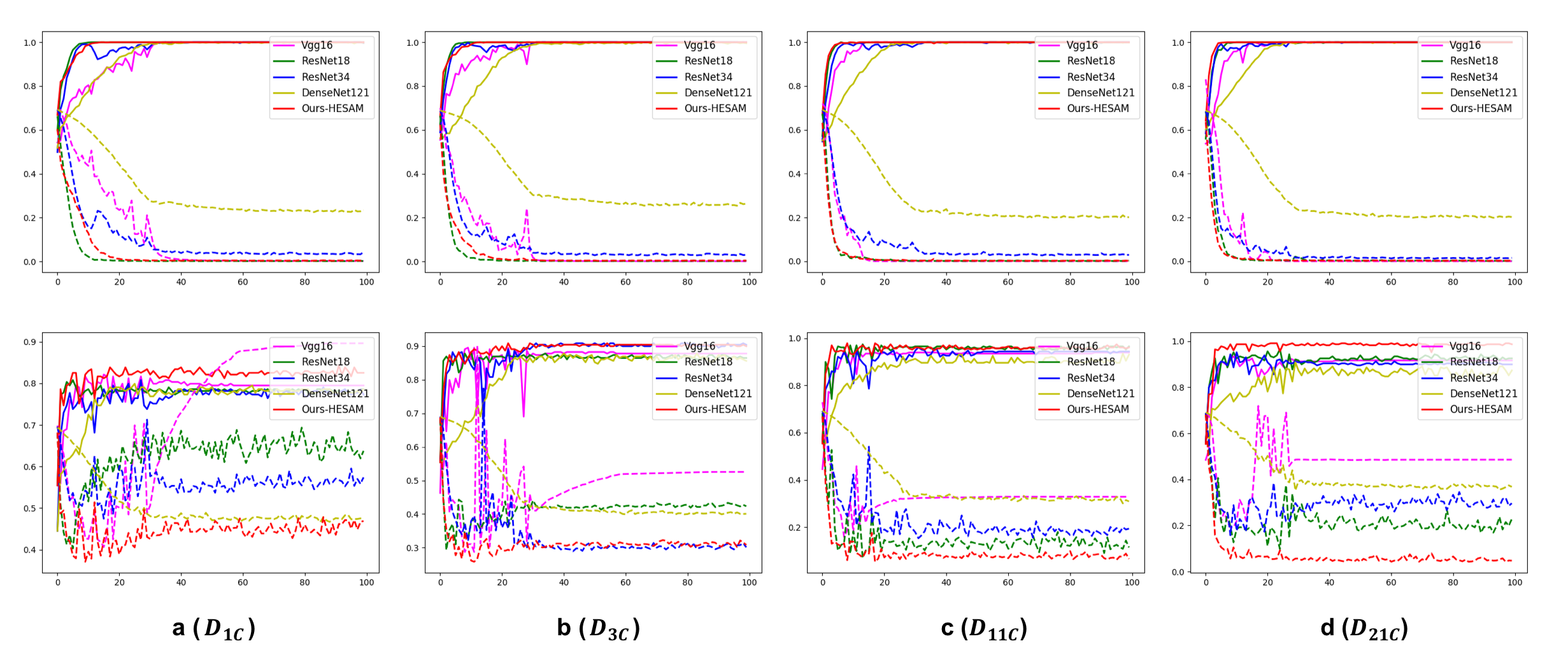}
  }
  \caption{Training and test performances on 4 constructed datasets, $D_{1C}$, $D_{3C}$, $D_{11C}$ and $D_{21C}$. Solid lines denote accuracies and dashed lines represent losses. Subfigures in the first and second row represent training and test results, respectively.}
  \label{Fig.curve}
\end{figure*}

\textbf{Alleviating overfitting through multi-slice input.} We use multi-slice data to instead manual data augmentations which are tailored for reducing overfitting. From the results shown in Fig. \ref{Fig.curve}, all the methods except for DenseNet121 suffer from overfitting, as shown in Fig. \ref{Fig.curve}a. With the increment in the number of input channels, the overfitting issue is gradually alleviated, as shown in Fig. \ref{Fig.curve}b and \ref{Fig.curve}c. Furthermore, it is observed in Fig. 3d that the test losses of all the methods got increased relative to Fig. \ref{Fig.curve}c, because nodule-irrelevant information occupies a larger proportion of the whole volume, rendering extracted features less discriminative. However, our HESAM method exhibits the least overfitting in Fig. \ref{Fig.curve}a, which can be effectively addressed with increasing the number of input channels. For VGG16, although the overfitting happens in all datasets, the test loss is decreased in fact and the corresponding test accuracies improve significantly with the increase of input channels.

DenseNet121 is the deepest structure in our experiments. Theoretically, it should learn more useful and accurate features. From all the subfigures in Fig. \ref{Fig.curve}, the performances of DenseNet-121 are not severely influenced by the number of channels, since it cannot be trained well on our datasets given relatively small amounts of data.  Although the first convolution layer is able to extract diverse and robust representation via multiple slices, this advantage will be weakened with the increase of depth of the model. Briefly speaking, under our experimental setting, the too deep network (DenseNet-121) could not perform better than shallower networks even utilizing multi-channel input because of the limited dataset.

\begin{figure}[ht]
  \centering
  \setlength{\abovecaptionskip}{0.5pt}
  \subfigure{
  \hspace{-0.4cm}
    \includegraphics[width = 5.5in]{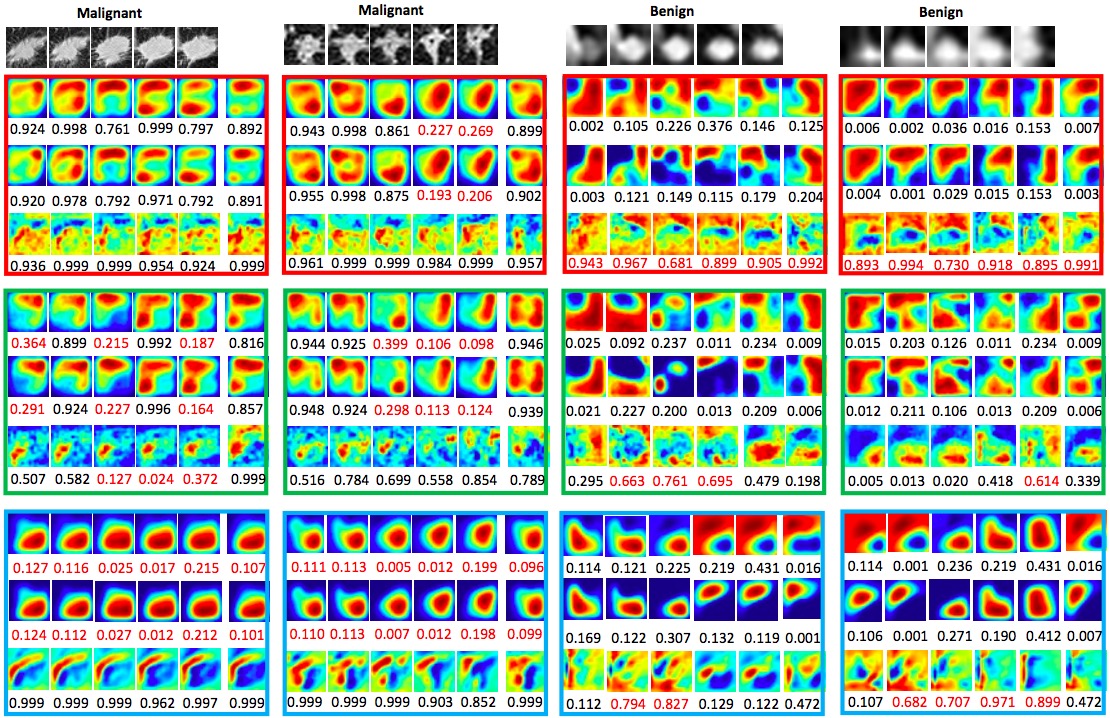}
  }
  \caption{Visualization of CAMs, Grad-CAMs, and SAMs obtained by models that are trained on $D_{11C}$. Each nodule is illustrated 5 of 11 slices. Activation maps at corresponding positions are obtained by feeding the volume of 11 channels (duplicated single slice) into the trained models. The final column of each color block shows the results obtained by feeding volume of all 11 adjacent slices. Note that the training data has the same number of channels with test data. \textcolor{red}{Red}, \textcolor{green}{green} and \textcolor{blue}{blue} block denotes the results generated by ResNet18-CAM/SAM, ResNet34-CAM/SAM, and VGG16-CAM/SAM, respectively. The first row, the second row and the third row of each color block show CAMs, Grad-CAMs, and SAMs, respectively. Values are predictions of the malignant category. Red color numerics represent false positive (FP) benign nodules and false negative (FN) malignant nodules.}
  \setlength{\belowcaptionskip}{-3cm}
  \label{cam_grad_sam}
\end{figure}

\subsection{Comparison Between CAM and SAM}
CAM is difficult to localize fine-grained and discrete attention regions. SAM solves this problem effectively by projecting final feature maps into corresponding weights in a soft manner. The weights are updated separately and SAM maintains specificity of each feature. We applied ResNet18-CAM/SAM, ResNet34-CAM/SAM and VGG16-CAM/SAM (Tab. \ref{results_1}) to demonstrate distinctions between CAM, Grad-CAM and SAM. Most of the methods in Tab. \ref{results_1} achieve better performances when using SAM, and this can further justify that SAM indeed pays more attention to distinctive LNSM regions than CAM and Grad-CAM. In addition, fine-grained regions are preserved which can be seen in Fig. \ref{cam_grad_sam}.

As illustrated in Fig. \ref{cam_grad_sam}, CAMs localize on malignant and benign nodules at a larger area, i.e., coarse regions of lesion. SAMs can localize on relatively fine-grained feature regions which are exhibited apparently discontinuous localizations of margins and micro-regions. However, SAMs also result in false negatives (FNs) and false positives (FPs), which will be avoided by HESAM. For one method, activation maps in Fig. \ref{cam_grad_sam} of every single slice are different from each other. This demonstrates that the models pay attention to different regions of each slice. The results of a whole volume cover attention regions in all single slices. That is why the performances of each model increase along with growing input channels (Tab. \ref{results_1}). Discriminative information in all slices makes great contributions to performance improvement. Each model learned different features for both categories of nodules and focused on varying regions. Most distinguishing features are variations in the margin of malignant nodules, whereas most benign nodules demonstrate smoother contour compared to malignant nodules. Note that, the high attention regions in activation maps denote where the model looks at during prediction rather than the concrete features.

\subsection{CAM, SAM and HESAM on Our Model}
This experiment verifies the effectiveness of high-level feature enhancing operation. 
\begin{figure}[ht]
  \centering
  \setlength{\abovecaptionskip}{0.5pt}
  \subfigure{
  \hspace{-0.4cm}
    \includegraphics[width = 6in]{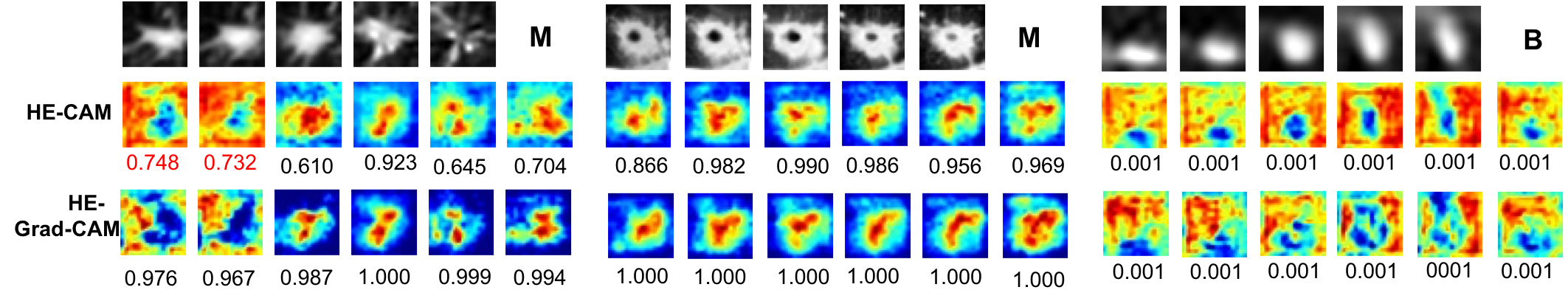}
  }
  \caption{Comparison of attention maps generated by the model w/o high-level feature. HE-. represents the maps are combined with high-level feature. The value under each map denotes the probability of being classified as the malignant class.}
  \setlength{\belowcaptionskip}{-3cm}
  \label{he_cam_grad_cam}
\end{figure}

\begin{figure*}[ht]
  \centering
  \setlength{\abovecaptionskip}{2pt}
  \subfigure{
    \includegraphics[width = 6in]{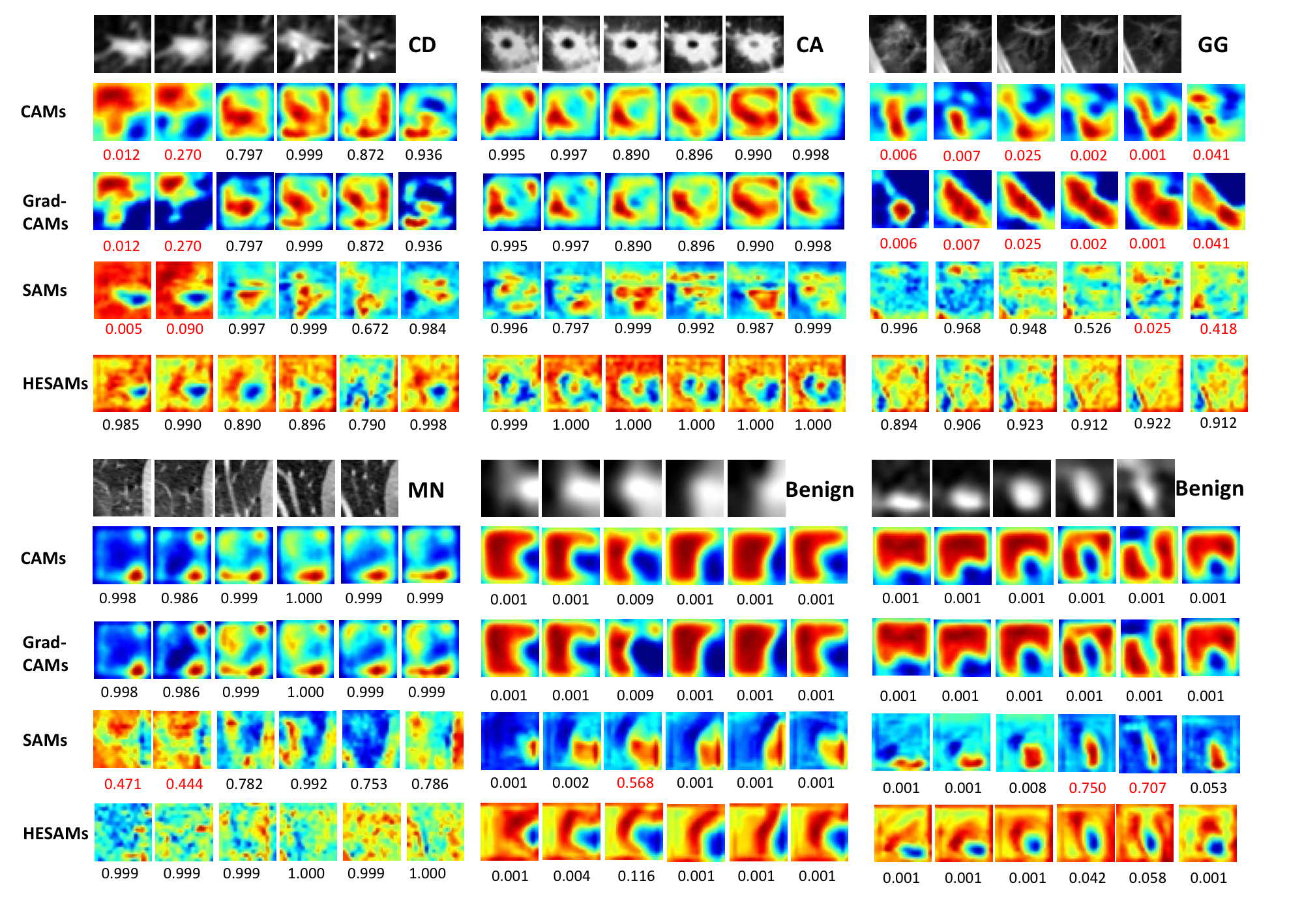}
  }
  \caption{CAMs, SAMs, and HESAMs of CD, CA, GG, MN, and two benign nodules, which are generated by 3 versions of our model trained on $D_{11C}$. Values under each map are predictions of the malignant category. Red color numerics represent false positive (FP) benign nodules and false negative (FN) malignant nodules negatives.}
  \label{hesam}
  \setlength{\belowcaptionskip}{0.5pt}
\end{figure*}

Maps in the same positions in Fig. \ref{hesam} have the same meaning as those in Fig. \ref{cam_grad_sam}. The CAM and SAM  versions of our model (Ours-CAM, Ours-SAM) have turned into using only final residual learning features to make a classification. As shown in Fig. \ref{hesam}, CAMs, and Grad-CAMs localize not only on some of the edges but also on part of nodule bodies. Consequently, this results in high FNs and FPs. SAMs performed as imagined to focus on margins and variations of shape, which is benefited from the discretization of local features preserved by average pooling without GAP and the corresponding separately updated parameters of multiple FC layers. Note that unlike CAM and Grad-CAM, FNs and FPs of SAM almost locate at random regions of margins rather than nodule bodies. The phenomenon above is caused by final residual learning features which are insufficient to learn more semantic information with respect to some special cases (such as shape variations and edge blurring related to nodule structures). In other words, these methods fail to highlight the most representative and semantic LNSM among final Conv features.

HESAM reapplies the high-level compressed information of individual nodule for localization and improves the classification accuracies simultaneously (Tab. \ref{results_1}). HESAM rectifies or augments the weights of discriminative LNSM features which are illustrated in the fifth row of each nodule in Fig. \ref{hesam}. HESAM and SAM have the same advantage that they can localize on discontinuous and more correct LNSM regions of malignant nodules, and focus on regions of smooth and larger area of benign nodules. The shortage of SAM is that it is not able to cover all discrete regions just as HESAM does. The advantage of localizing on fine-grained morphology variations of HESAM is apparently displayed in CD, CA, MN and GG (Fig. \ref{hesam} and \textit{Fig. A.2 in Appendix A}). The irregular spiculation and cavity are localized completely. For GG and MN, CAM and SAM prefer to search the area where the variations are relatively smooth. In order to further explore the effectiveness of high-level features, we have trained a new model using our implementation that contains GMP (high-level feature), RB and GAP. From Figs. \ref{he_cam_grad_cam} and \ref{hesam} we can see that with the combination of high-level feature, the obtained HE-CAMs and HE-Grad-CAMs cover the similar highlighted regions to their counterparts those without high-level feature. The big difference is that HE-CAMs and HE-Grad-CAMs are more discrete, and this attributes to the enhancement of a part of final features.

From Tab. \ref{results_1}, we can see that HESAM method exhibits outstanding and robust performances on all datasets and this can be closely linked with learned LNSM features (see \textit{Fig. D.5, Appendix D}). And all the experiments in Fig. \ref{cam_grad_sam} and Fig. \ref{hesam} illustrate that volumes consisting of same single slices are more likely to be misclassified and mislocalized when using CAM, Grad-CAM and SAM. However, HESAM performs better even on volumes of single slices. Therefore, we demonstrate that 2D convolutions on transverse planes of all slices extract various LNSM features for one nodule, and this enables the model to make decisions based more on margin variations than nodule size which is a fuzzy descriptor under different scales.

\subsection{Ablation Study}
In this section, we will discuss three aspects of our implementation: (1) the effects of different global pooling modes (GAP and GMP) applied to high-level features (HF) and final Conv features (FCF) in Fig. \ref{net}; (2) conducting ``sum'' or ``concatenation'' to fuse high-level feature vector and weights of final Conv features in Fig. \ref{net}; (3) effectiveness of different features in our network, i.e., features yielded by GMP and residual blocks.

\subsubsection{GAP or GMP}

As stated in \citep{Zhou_2016_CVPR}, GAP encourages a network to identify the extent of the object and GMP encourages it to identify just one discriminative part, and this has been verified experimentally on ILSVRC \citep{ILSVRC}. In order to verify how the different pooling modes affect our method, we have conducted the experiments with varying settings of pooling modes for HF and FCF on $D_{11C}$ dataset. Here, changing the pooling mode for FCF equals to changing the pooling mode in SAM structure. The classification results and visualizations of activations are shown in Tab. \ref{gap_gmp_acc} and Fig. \ref{gap_gmp_fig}, respectively.
\begin{table}[H]
  \centering
  \setlength{\belowcaptionskip}{5pt}
  \caption{Performances of our model ($D_{11C}$ dataset) using different settings of pooling modes for high-level features (HF) and final Conv features (FCF).}
    \begin{tabular}{p{3cm}<{\centering}p{2cm}<{\centering}p{2cm}<{\centering}p{2cm}<{\centering}}
      \hline
       Settings (HF / FCF) & Accuracy & Sensitivity & Specificity\\
      \hline
       GAP / GAP & 96.51 & 0.9216 & 1.0 \\
       GAP / GMP & 96.51 & 0.9216 & 1.0 \\
       GMP / GMP & 96.94 & 0.9313 & 1.0 \\
       GMP / GAP & 99.13 & 0.9705 & 0.9921\\
      \hline
    \end{tabular}
  \label{gap_gmp_acc}
\end{table}

Stretching back to our scheme, SAM is tailored for discretization of discriminative features in final Conv layer and high-level feature enhancement aims to highlight the nodule structure during computing the activation maps. 

Now, let us consider the HF part. HF preserves the extent structure of nodule volume (see \textit{(a) and (c) in Fig. A.1, Appendix A}), even the HF of different kinds of nodules are not discriminative in visualization. Therefore, the HF vector (after global pooling) of $256$ dimensions is not used as the input of classifier directly. Through the element-wise sum of HF vector and final feature vector resulted by SAM, the weights of some final Conv features are refined when we compute the HESAMs, i.e., HF vector plays a role of highlighting some final Conv features based on nodule structure. Then, different pooling modes applied to HF result in different enhancements of features. GAP produces an HF vector with more similar elements, and it will give almost the same highlights to final Conv features, i.e., features (discriminative or non-discriminative) contributed equally to HESAMs. From Fig. \ref{gap_gmp_fig}, we can see that HF with GAP brings larger attention regions around the nodule bodies. In contrast, GMP produces an HF vector that contains elements with greatly different values, thus there will be apparent distinctions among weights of final features. And the fourth and fifth rows in Fig. \ref{gap_gmp_fig} illustrate fine-grained attention regions.

In addition, different pooling modes for FCF, i.e., pooling in SAM, also can result in varying classification performance and attention regions. If we conduct GMP on SAM, each local feature after GMP represents only one discriminative part of the corresponding final feature. High attention regions generated by applying GMP on SAM are more discrete than those of using GAP (see Fig. \ref{gap_gmp_fig}, the third row and fourth row are more discrete than the second row and fifth row respectively). For the case of GMP / GMP setting, the attention regions seem randomly mapping. This may because the HF vector and weights obtained by SAM are all consisted of many distinctive values, and the sum of them will result in a sparse vector. Therefore, a few final features are equipped with higher weights. Moreover, we have examined the experiments that rolling the elements ($50$ one time) in HF vector. These experiments aim to obtain the summed vector through element-wise sum at different positions. And the attention maps for GMP / GMP are also random. 

\begin{figure*}
  \centering
  \setlength{\abovecaptionskip}{2pt}
  \subfigure{
    \includegraphics[width = 5.6in]{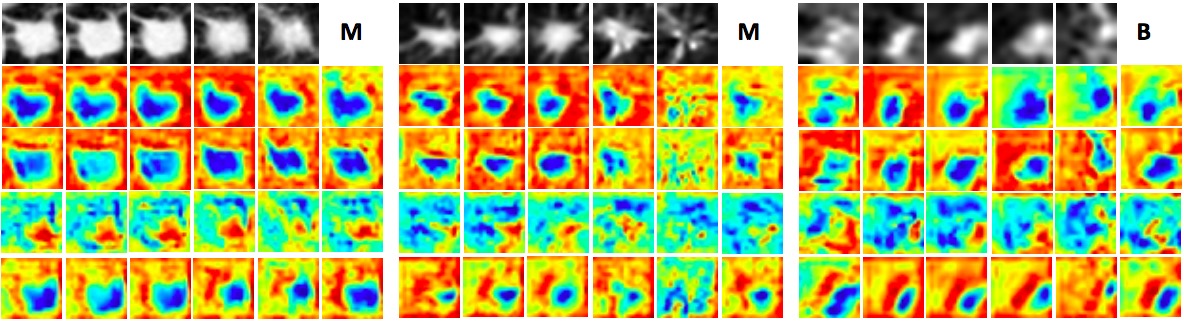}
  }
  \caption{HESAMs generated by our model with three settings of pooling modes for HF and FCF (from the second row to the fitth): GAP / GAP, GAP / GMP, GMP / GMP and our best results of GMP / GAP. M and B represent malignant and benign, respectively.}
  \label{gap_gmp_fig}
  \setlength{\belowcaptionskip}{0.5pt}
\end{figure*}
\begin{figure*}[ht]
  \centering
  \setlength{\abovecaptionskip}{2pt}
  \subfigure{
    \includegraphics[width = 5.8in]{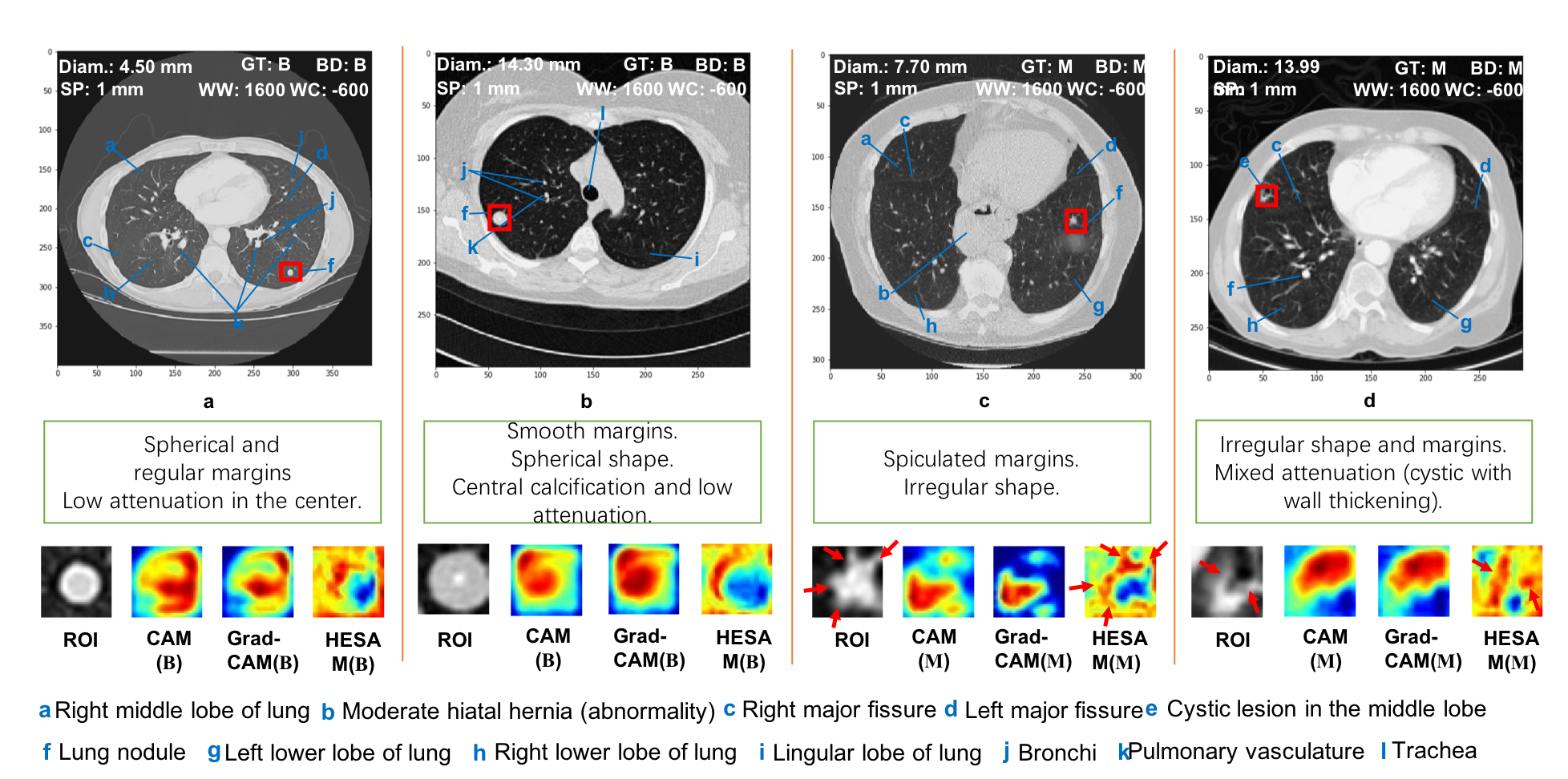}
  }
  \caption{Visually matching results via radiologists study in full-size CT slices. \textcolor{green}{Green} boxes contain  LNSM descriptions of nodules given by radiologists. Nodule diameter (Diam.), ground truth (GT) and blinded decisions (BD) are provided at upper of each slice. Nodules are bounded manually (\textcolor{red}{Red} boxes) according to positions of nodule centers and diameters in Dicom files. Parentheses contain classification results of our model equipped with different mapping methods, ${\bf M}$ and ${\bf B}$ represent malignant and benign respectively. \textcolor{red}{Red} arrows overlapped on malignant ROIs and corresponding HESAMs indicate the discriminant regions.}
  \label{analysis_fig}
  \setlength{\belowcaptionskip}{0.5pt}
\end{figure*}

\subsubsection{Sum or Concatenation for Fusion of HF Vector and Weights of FCF}
In our model, we use element-wise sum to highlight the weights of corresponding final features in the computation of HESAMs. If we concatenate HF vector and the output of SAM at channel dimension in practice which is the same as \citep{pn-samp}, there will be a final feature vector of $512$ dimension. So the number of parameters is increased, and the best classification accuracy on $D_{11C}$ is only $97.83\%$. It is not difficult to give the reason of this degradation that high-level features of both kinds of nodules (\textit{Fig. A.1, Appendix A}) are not discriminative enough even if high-level features are more semantic. Most importantly, the final feature vector of $512$ dimension will lose the activation mapping function, because the FCF composes $256$ feature maps and this number is not equal to $512$.

\subsubsection{Effectiveness of Different Features}
In this ablation experiment, we explore the effectiveness of features generated by GMP and residual blocks. In simplicity, we use F1 and F2 to denote them, respectively.

As can be seen in Tab. \ref{gmp_rb_ablation}, we use U+SAM without RB and GMP as a baseline. The methods U+GMP+SAM and U+RB+SAM correspond to F2 and F1, respectively. For fair comparisons, the number of output channel in U structure is set as 256 for all methods in Tab. \ref{gmp_rb_ablation}, because it is needed in the high-level enhancement scheme using U+GMP+SAM. The accuracy of U+RB+SAM is higher than that of U+GMP+SAM, indicating that features yielded by RB are more discriminative than low-resolution features generated by GMP. An interesting observation is that the sensitivities of U+RB+SAM are higher than those of U+GMP+SAM on all datasets. Therefore, F2 is prone to identifying shapes and margins of malignant nodules. While U+GMP+SAM method has the competitive specificities compared with U+RB+SAM. Especially in $D_{1C}$ case, U+GMP+SAM achieves the highest specificity. This helps our model reduce the false positive rate.

\begin{table}[h]
 \centering
  \setlength{\belowcaptionskip}{5pt}
  \caption{Comparisons of different features in our architecture. U denotes the UNet structure in the proposed method, RB is the abbreviation of residual blocks. These methods are trained on our four constructed datasets.}

    \begin{tabular}{p{2cm}|p{4cm}p{1.5cm}<{\centering}p{1.5cm}<{\centering}p{1.5cm}<{\centering}p{1.5cm}<{\centering}}
      \hline
       Metrics & Methods & $D_{1C}$ & $D_{3C}$ & $D_{11C}$ & $D_{21C}$ \\
      \hline
                & U+SAM (Baseline) & 82.53 & 89.96 & 96.94 & 96.51 \\
       Accuracy & U+RB+SAM 		   & \textcolor[rgb]{0,0,1}{84.28} & \textcolor[rgb]{0,0,1}{91.27} & 		 \textcolor[rgb]{0,0,1}{98.69} & \textcolor[rgb]{0,0,1}{98.25} \\
                & U+GMP+SAM 	   & 83.41 & 90.39 & 98.25 & 96.94 \\
                & U+GMP+RB+SAM 	   & \textcolor[rgb]{1,0,0}{84.72} & \textcolor[rgb]{1,0,0}{93.01} & \textcolor[rgb]{1,0,0}{99.13} & \textcolor[rgb]{1,0,0}{98.69} \\
      \hline
                   & U+SAM (Baseline) & 0.7353 & 0.8627 & 0.9411 & 0.9314 \\
       Sensitivity & U+RB+SAM 		  & 0.8137 & 0.8725 & 0.9608 & 0.9706 \\
                   & U+GMP+SAM 		  & 0.7059 & 0.8529 & 0.9314 & 0.9412 \\
                   & U+GMP+RB+SAM 	  & 0.7549 & 0.8921 & 0.9705 & 0.9706 \\
      \hline
                   & U+SAM (Baseline) & 0.8976 & 0.9291 & 0.9921 & 0.9921 \\
       Specificity & U+RB+SAM 		  & 0.8661 & 0.9449 & 1.0 & 0.9921 \\
                   & U+GMP+SAM 		  & 0.9370 & 0.9449 & 1.0 & 0.9921 \\
                   & U+GMP+RB+SAM 	  & 0.9212 & 0.9606 & 0.9921 & 1.0 \\
      \hline
    \end{tabular}
  \label{gmp_rb_ablation}
\end{table}

\subsection{Visually Matching Experiment}
It is known that activation maps are difficult to quantitatively measure which mapping method covers more discriminative and critical regions \citep{Zhou_2016_CVPR,gradcam}. Therefore, we conduct the matching experiment incorporated with radiologists' blinded decision process. That is to say, we treat the "radiologists' descriptions" as the pseudo standard quantitation to measure the activation maps. 

\textbf{Blinded decision}. To avoid the effect of nodule size on the blinded decision, we select 20 nodules of both kinds that are in similar sizes (some special cases like benign nodules with larger size and malignant nodules with smaller size are included, and Fig. \ref{analysis_fig}, shows four examples of matching). During the process of blinded decision, we provide radiologists with only full-size CT slices and bounded nodules in the corresponding slices. And the radiologists are required to give the categorization decisions and LNSM descriptions of bounded nodules.

The obtained blinded decision and description results are used to match the discriminative regions covered by activation maps of different methods. As shown in Fig. \ref{analysis_fig}, $(a)$ and $(b)$ are benign nodules and $(c)$ and $(d)$ are malignant nodules. Meanwhile, $(a)$ and $(c)$ are with smaller size, and $(b)$ and $(d)$ are with larger size. According to blinded descriptions (\textcolor{green}{green} boxes), benign nodules possess spherical shape and regular/smooth margins, which are better covered by highlighted regions of HESAM compared with CAM and Grad-CAM. In another hand, blinded descriptions of malignant nodules are with spiculated/irregular margins and irregular shapes, and our HESAM also pays more attention to these discrete and fine-grained LNSM features. However, CAM and Grad-CAM are not equipped with this advantage and they localize more on nodule bodies or small parts of irregular margins and shapes, which are biased to radiologists' blinded descriptions. In other words, the attention regions of CAM and Grad-CAM do not match the margins and shapes where the radiologists care about. Simultaneously, all the classification results of different mapping methods are consistent with GT and BD. Consequently, HESAM indeed makes the correct predictions based on the discriminative LNSM features which are critical to clinical diagnosis.

\subsection{Significance Test}
We have conducted the Wilcoxon signed-rank test \citep{wilcoxon} for the prediction probabilities of test images. Tab. \ref{sig_analysis} compares $p$-values in the Wilcoxon signed-rank test between predictions of our HESAM method and those of other methods. All the $p$-values being less than 0.01 suggest that Ours-HESAM is significantly different from others for the prediction. That is to say, the output space of test set spanned by our method is significantly different and more separable than that by the other methods.
\begin{table}[ht]
  \centering
  \setlength{\belowcaptionskip}{5pt}
  \caption{$p$-values in the significant tests between Ours-HESAM method and other methods trained with different constructed datasets.}

    \begin{tabular}{p{4.8cm}p{2cm}<{\centering}p{2cm}<{\centering}p{2cm}<{\centering}p{2cm}<{\centering}}
      \hline
       Methods & $D_{1C}$ & $D_{3C}$ & $D_{11C}$ & $D_{21C}$ \\
      \hline
       PN-SAMP \citep{pn-samp}              & <0.01 & <0.01 & <0.01 & <0.01 \\
       SAG-C \citep{schlemper2019attention} & <0.01 & <0.01 & <0.01 & <0.01 \\
       ResNet18 \citep{ResNet}              & <0.01 & <0.01 & <0.01 & <0.01 \\
       ResNet34 \citep{ResNet}              & <0.01 & <0.01 & <0.01 & <0.01 \\
       VGG16 \citep{vgg}                    & <0.01 & <0.01 & <0.01 & <0.01 \\
       DenseNet121 \citep{densenet}         & <0.01 & <0.01 & <0.01 & <0.01 \\
      \hline
    \end{tabular}
  \label{sig_analysis}
\end{table}

\subsection{Inter-annotator Variability Study}
Because LIDC dataset contains some significant inter-annotator variability, we applied 20 nodules (10 malignant and 10 benign) in a {\bf Blinded decision} experiment to verify the merits of our network. {\it These nodules are accurately classified by our model ($100\%$ accuracy, $0$ FPs and $0$ FNs) and the predicted labels are consistent with the judgments by the involved radiologists}. These 20 nodules were annotated by at least 3 doctors, and the evaluation results are shown in Tab. \ref{doc_analysis}. The accuracy of each doctor was calculated by ignoring the nodules  with label 3 (unlabeled nodules were excluded in the evaluation of this doctor). Also, no attention was paid  on whether the nodule labeled 3 is a false positive or a false negative.

Tab. \ref{doc_analysis} shows that 4 radiologists performed significantly different in terms of FPs and FNs, and our method is able to reduce the FPs and FNs and outperforms average radiologists.

\begin{table}[ht]
  \centering
  \setlength{\belowcaptionskip}{5pt}
  \caption{Separate evaluations with respect to different doctors. FP and FN denotes false positive and false negative respectively.}

    \begin{tabular}{p{3cm}|p{2cm}<{\centering}p{2cm}<{\centering}p{2cm}<{\centering}p{2cm}<{\centering}}
      \hline
       Doctors       & Dr. 1  & Dr. 2  & Dr. 3  & Dr. 4 \\
      \hline
      \hline
       Accuracy (\%) & 88.26 & 82.35 & 94.44 & 100 \\
      \hline
       No. of FPs    & 1 & 2 & 1 & 0 \\
      \hline
       No. of FNs    & 1 & 1 & 0 & 0 \\
      \hline
    \end{tabular}
  \label{doc_analysis}
\end{table}
  
\section{Discussion}
A number of deep learning methods for medical tasks lack applicability that is ascribed to many kinds of data preprocessing procedures of medical volume and inflexibility of deep architectures. Consequently, deep medical imaging tasks are very different from image classification / detection on ImageNet or COCO \citep{ILSVRC, vgg, densenet, ResNet, alexnet, googlenet}. Multi-view CNN \citep{2016multiview} applied the input data of cropped CT patches of nine different views centered by volume center, and this results in a complicated model which is difficult to extend to other studies. TumorNet \citep{2017tumornet} also would like to use patches of one volume at different views, and the generated patches through median intensity projection seem to be a subset of those in \citep{2016multiview}. Multi-scale patches \citep{multiscale_cnn} achieve less increase in nodule categorization performance, since the extracted shape features, which should have been sensitive to CNN, are changeless with respect to one nodule (patch). \citep{deeplung, pn-samp, 2017multilevel3d} utilized 3D architectures to make classifications between nodule and non-nodule, malignant and benign. However, the sizes of input data and pixel-spacing value in their studies are quite different. Moreover, the nodule categorization performance is conditioned on the power of nodule detection model in DeepLung. 

For the lack of annotated data in medical cases, transfer learning \citep{how_trans, hoo2016deep, trans_survey} have been introduced to solve this problem significantly. Most of transfer learning approaches are based on models that pre-trained on ImageNet, then they initialize the model via weights of pre-trained models or utilize the off-the-shelf features directly. Nevertheless, the number of channels of input data for pre-trained models must be the same as those in the first layer of corresponding models, e.g., \citep{fully3d2017, hoo2016deep, shan20183d} duplicated the nodule patches and resulted in the 3-channel input, and this is what we have done for testing classification and localization performances under single slice in our experiments. We use a relatively simple preprocessing method, i.e., extracting adjacent slices. In addition, various shapes and margins of different slices can be captured by our 2D model. Moreover, the number of parameters in our model is less than that of in 3D models, so that we have encountered fewer difficulties during the training phase.

\section{Conclusion}
Our paper proposes a novel method for fine-grained lung nodule shape and margins localization through soft activation mapping (SAM). Our method provides an accurate interpretation of lung nodule categorization and better classification performance coupled with high-level features from the classification loss of the UNet model. Testing on the LIDC-IDRI dataset suggests that feeding multiple slices into 2D networks can also capture the discriminative information in each slice and alleviate the overfitting problem. Our model has outperformed representative state-of-the-art models in terms of false positive rate. However, our model depends on the extraction of high-level features from an auto-encoder-based architecture, and dimensions of high-level features and final weights must be the same. 

In future, we will extend SAM to a wider range of applications in the medical domain such as lesion detection (testifying where the network pays attention during nodule detection within the whole CT slices) and apply more adaptive methods for further improvement of the lung LDCT screening and diagnostic performance.

\section*{Acknowledgments}
This work was supported in part by the Shanghai Municipal Science and Technology Major Project (No.2018SHZDZX01) and ZJLab, National Natural Science Foundation of China (NSFC 61673118), the National Key R \& D Program of China (No. 2018YFB1305104) and NIH U01 EB017140.

\section*{References}

\bibliography{references}

\end{document}